\documentclass[conference]{IEEEtran}
\usepackage{cite}
\usepackage{amsmath,amssymb,amsfonts}
\usepackage{algorithmic}
\usepackage{graphicx}
\usepackage{textcomp}
\usepackage{mathtools}
\usepackage{xcolor}
\usepackage{gensymb}

\def\BibTeX{{\rm B\kern-.05em{\sc i\kern-.025em b}\kern-.08em
    T\kern-.1667em\lower.7ex\hbox{E}\kern-.125emX}}
\begin{document}

\title{Open-endedness induced through a predator-prey scenario using modular robots
}

\author{\IEEEauthorblockN{1\textsuperscript{st} Dimitri Kachler}
\IEEEauthorblockA{\textit{Computer Science dept.} \\
\textit{Vrije Universiteit Amsterdam}\\
Amsterdam, Netherlands \\
d.r.kachler@student.vu.nl}
\and
\IEEEauthorblockN{2\textsuperscript{nd} Karine Miras}
\IEEEauthorblockA{\textit{Computer Science dept.} \\
\textit{Vrije Universiteit Amsterdam}\\
Amsterdam, Netherlands \\
k.dasilvamirasdearaujo@vu.nl}
}

\maketitle

\begin{abstract}

This work investigates how a predator-prey scenario
can induce the emergence of Open-Ended Evolution (OEE). We
utilize modular robots of fixed morphologies whose controllers
are subject to evolution. In both species, robots can send
and receive signals and perceive the relative positions of other
robots in the environment. Specifically, we introduce a feature we call a \textit{tagging system}: it modifies how individuals can perceive each other and is expected to increase behavioral complexity. Our results show the emergence of adaptive strategies, demonstrating the viability of inducing OEE through predator-prey dynamics using modular robots. Such emergence, nevertheless, seemed to depend on conditioning reproduction to an explicit behavioral criterion.

\end{abstract}


\section{Introduction}
The longest evolutionary experiment has been continually running on planet Earth for the past 3.7 billion years~\cite{PBSEvolution}: natural life. Throughout this long period, organisms have only been preying on each other for the last 1.2 billion years. Evolutionary Computation (EC), on the other hand, has only existed for the last 70 years~\cite{EvolutionHist} and has spawned a variety of different approaches. Nevertheless, the dominant paradigm of EC has been the inversion of the concept of fitness from a metric measured \textit{a posteriori} to a metric measured \textit{a priori}: from being considered fit in case your traits allow you to survive and reproduce to being allowed to survive and reproduce in case you possess certain traits.

While effective in diverse domains, this paradigm is limited because it lacks crucial aspects that would allow the emergence of complexity~\cite{LISA}: not all beneficial processes translate into a numerical gain or are adequately represented by a singular scalar value. 

 To address the aforementioned challenges, a different paradigm has been explored in the literature, which is closer to natural evolution: Open-Ended Evolution (OEE)~\cite{OEEGrand,taylor2016open}. Two core axioms unique to Open-Ended Evolution state that concepts of fitness and generations are applied implicitly rather than explicitly~\cite{Towards}. Firstly, there is no actual fitness function to judge an individual through selection. As a result, the selection process for inheriting genes must not directly discriminate against solutions; discrimination may only arise through indirect organic means - mechanisms that arise as a result of the system dynamics.
 
 Specifically within EC, attempts at OEE started with the Artificial Life (ALife) community, using artificial worlds such as Tierra~\cite{Tierra}, where programs could self-replicate and compete for computation power and memory space. Another example was Polyworld~\cite{Polyworld}, where agents could eat or mate with each other. Polyworld exhibited predator-prey dynamics with open-ended characteristics, but the underlying physical representation for the agent bodies was 2D polygons. Beyond 2D worlds, OEE has also been applied to wheeled robots in hardware~\cite{EDEA} and even to
  modular robots~\cite{Swarm}, which are more challenging to work with than wheeled ones. However, we are unaware of any work combining modular robots and predator-prey dynamics in the context of OEE.
 
%considering that there are no explicit selection mechanisms,

Therefore, the present work investigates how a predator-prey scenario can induce the emergence of OEE. Specifically, we introduce a novel systemic feature that we call \textit{tagging system}: this system modifies how individuals can perceive each other and is expected to promote behavioral complexity. 

Because the need for a \textit{minimum criterion} has been suggested before in the literature~\cite{LISA}, we hypothesize that: \textit{in the current predator-prey scenario, the emergence of OEE depends on the existence of an explicit behavioral criterion to allow reproduction}.

 \section{Methodology}

The code to reproduce all experiments is available on GitHub\footnote{https://github.com/NanoNero1/revolve2-multi}. All experiments were repeated 10 times for statistical significance. All parameters were chosen empirically. Each experiment was run for 6000 seconds with a timestep of 0.0012. A video showing robots during one of the experiments is available\footnote{https://youtu.be/czv-cwoAk0g}.

\subsection{Robot Body and Environment}
 \begin{figure}[htbp]
\centerline{\includegraphics[scale=0.5]{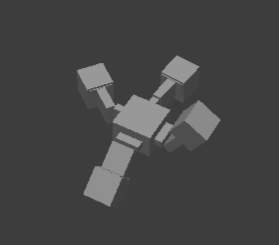}}
\caption{A \textit{spider-shaped} robot body simulated in Mujoco.}
\label{spier}
\end{figure}

The experiments are simulated using the Mujoco physics engine, wrapped by a robot framework called Revolve~\cite{revolve, hupkes2018revolve}. The robot bodies are modular bodies~\cite{auerbach2014robogen} constructed by connecting blocks and joints (motors). For an initial proof of concept, the current experiments utilize a fixed body configuration (Fig. \ref{spier}), but the long-term view of future work envisions allowing these configurations to evolve. In all experiments, the robots evolve in a square, flat plane surrounded by walls. % A unit means a distance of length $x$ on the ground. 
 % 20 units by 20 units 

\subsection{Robot Controller}

The robot controller comprises two components: the \textit{targeted steering} and the \textit{cognitive brain}. The targeted steering influences the active behavior of the robot by controlling the motors. The cognitive brain influences both active and passive behaviors: it steers the gait that is generated with the targeted steering (active) and also changes a phenotypic trait that does not directly affect robot behavior (passive).

\subsubsection{Targeted Steering}

The targeted steering controller allows a robot to locomote to a specific target. In this case, one single controller was evolved in pre-experiments, and every robot received one independent copy of it. 

 This controller is a combination of a Central Pattern Generator (CPG)~\cite{CPG} with a steering mechanism that adjusts the outputs of the CPG. CPGs are networks capable of producing coordinated rhythmic activity patterns without sensory feedback inputs. Given a timestep, the CPGs generate values used to set the rotations of the motors.
 
  The usual approach in studies with predator-prey dynamics is to give predators direct access to the location of their closest prey and vice versa~\cite{CoEvolving, UVApreypred}. In opposition, we provide individuals with only the relative angle of a nearby individual of the opposite species. The steering mechanism~\cite{Targeted} adjusts the rotation of certain motors initially produced by the CPG using a target angle $\alpha$ derived from this relative angle - this $\alpha$ regards where the robot `wants' to go. For example, if a robot has a positive $\alpha$, it wants to go to the right. Thus, the robot should slow down motors on its right side by a scaling factor $\delta$. In the current experiments, $\delta$ is calculated with Eq.~\ref{eqdelta}, and the $\alpha$ is generated by the cognitive brain. To define what is left and right, an axis of symmetry is drawn diagonally (45\degree ) through the robot head. The determination of which side is designated as right or left adheres to a predefined frame of reference that establishes the front orientation.

\begin{equation}
\delta = (\frac{\pi - |\alpha|}{\pi})^2 \label{eqdelta}
\end{equation}

The values of the CPG parameters were produced with Compositional Pattern Producing Networks (CPPNs)~\cite{stanley2007compositional, stanley2009hypercube} evolved using the following parameters:
20 generations, 20 parents, 20 children, 20 population size, round-robin selection tournament, and 50 seconds of simulation time. The fitness function was the displacement towards the negative x-axis.

\subsubsection{Tagging System} 
We introduce a tagging system that limits the ability of robots to perceive other robots. Individuals may only perceive each other if they have the same tag. One useful analogy for the concept of a tag is that an individual may change passive phenotypic traits perceivable by their adversary, e.g., change their skin color to a color visible or invisible to the eyes of the adversary.

The tagging system allows each individual to choose their tag from either -1 or 1. To avoid erratic behavior, there is a cool-down of 50 seconds before a robot may switch its tag again. The motivation behind this system is to add complexity to the hunting process: it introduces a challenge to both predators and prey. For example, it may create situations like such: a predator is hunting down a prey, but mid-chase, the prey changes tag, rendering itself invisible to the predator. 

\subsubsection{Cognitive Brain}

\begin{figure}[htbp]
\centerline{\includegraphics[scale=0.28]{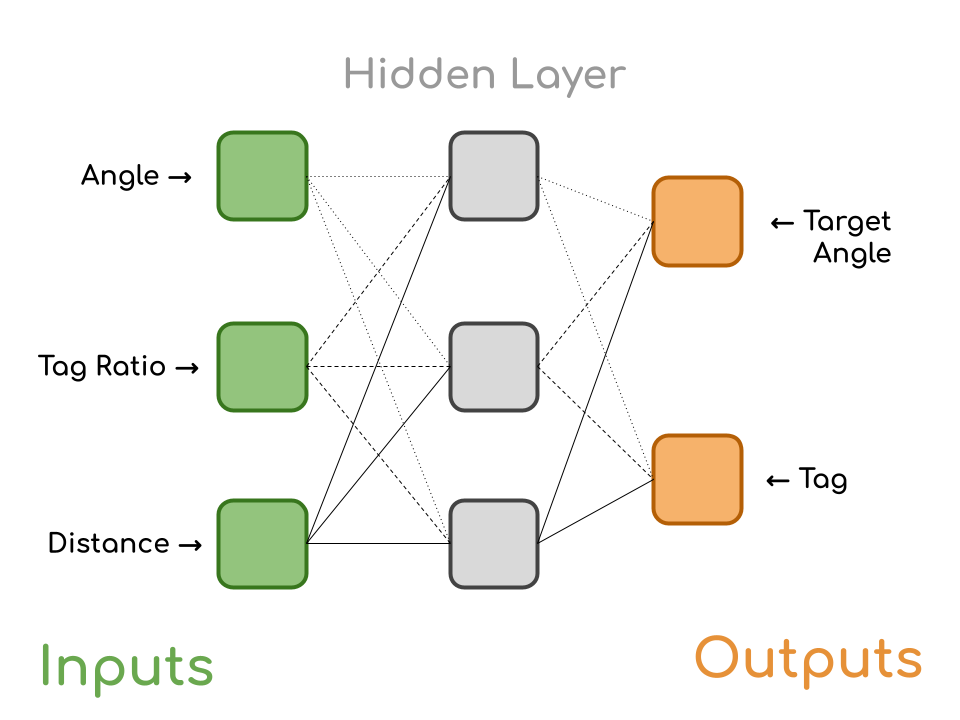}}
\caption{Architecture of the Cognitive Brain: a fully connected network.}
\label{cogb}
\end{figure}

The cognitive brain is a Fully Connected Neural Network that decides the tag of the robot and the target angle to use with the targeted steering (Fig. \ref{cogb}). The activation function utilized is the hyperbolic tangent function.  

%\begin{equation}
%\sigma (x) = \frac{2}{1+e^{-2x}}-1 \label{eqact}
%\end{equation}

 \paragraph{Inputs}
All inputs are bound inside [-1, 1]. Inputs \textit{Angle} and \textit{Distance} concern the closest adversary of an individual, whereas the input \textit{Tag Ratio} concerns the population. The term adversary will be re-occurring in later sections, and we define it as the closest observable robot of the opposite species. Therefore, the adversary of a prey is the closest predator within the same tag. There might be a predator even closer, but on a different tag: this is not the adversary.

The \textit{Angle} input is a value set to -1 if the adversary is on the left side of the robot and 1 if it is on the right.

The \textit{Distance} input is the distance to the adversary divided by the maximum terrain bounds.

 The \textit{Tag Ratio} input is defined with Eq. \ref{tagr} and calculates the ratio of how many robots are tagged as 1 relative to the total amount of robots. This variable informs individuals about the balance among the different tags within the population. We anticipate that this can be beneficial for a robot in making adaptive decisions regarding when to alter its tag. For instance, knowing that an excessive number of individuals share the same tag holds significance for a predator, as this disparity might indicate an overabundance of other predators within the same tag, resulting in heightened competition.

\begin{equation}
TR = \frac{P - \frac{N}{2} }{N}
 \label{tagr}
\end{equation}

\paragraph{Outputs}
The \textit{Target Angle} output is set to 0.7 radians if its output neuron is positive and -0.7 radians if it is negative - this value is provided to the targeted steering. The \textit{Tag} output is set to 1 if its output neuron is positive and -1 if it is negative.

 Furthermore, the cognitive brain is not queried for outputs at every timestep but every 2 seconds. Smaller intervals caused angle switches to be too erratic, while the chosen value produced smoother locomotion.

\subsection{Birth}

There is a fixed number of robot bodies in the environment: thirty robots. At birth, a new controller is attributed to a robot body already in the environment: this is possible when the controller previously inhabiting that body dies. Robots can be born in different ways, as described below.

%Additionally, the presence of a robot in the environment after birth does not mean it can immediately interact with other robots. Specifically, when prey are newly born, they are always invisible to everyone and immune to getting caught: this is true until they move five units away from any predator. 

\paragraph{Initialization}

When the experiment starts, 30 cognitive brain controllers are initialized with entirely randomized weights between -1 and 1 drawn from a uniform distribution. Each controller is attributed to one of the available bodies: 16 are prey, and 14 are predators.

\paragraph{Reproduction}
 
There are two forms of creating new controllers. First, there is a 1/3 chance of creating random controllers. This introduces new solutions to the gene pool, improving diversity. Second, there is a 2/3 chance of a new controller being implicitly sourced from an existing genotype. Implicit means that we do not use any explicit fitness function to evaluate individuals. When a predator catches a prey, the predator reproduces: the prey dies, and the offspring takes over the body avatar of the prey. As for prey reproduction, it happens when a predator dies: if any prey is within a certain minimum distance away from the predator, the closest prey to this predator reproduces. Similarly, the offspring of the prey takes over the body avatar of the predator. The offspring resulting from reproduction undergoes mutation by perturbing the network weights with values drawn from a normal distribution. 

%As for prey reproduction, it happens when a predator dies, and a prey is within four units away from it

\subsection{Death}

%To avoid `spawn-killing', a newborn prey must first move five units away

\paragraph{Prey Death}
A prey dies when a predator catches it. This happens when they find themselves within one unit from each other, regardless of whether they are on the same tag (prey is caught despite not being seen). To avoid `spawn-killing', a newborn prey must first move a certain minimum distance away from any predators before it becomes active and is eligible to be caught. Before this condition is met, the prey wanders around the map in a state of inactivity and is invisible to all predators. By `spawn-killing' we mean that the prey might have been born and placed in the environment (spawned) too close to predators. Additionally, there is an alternative mechanism by which prey may die: if the number of predators is nearing extinction, i.e., less than 7 predators, a prey is chosen to be sacrificed entirely at random. Similarly, no prey can die if there are only 7 prey in the population. These two constraints guarantee none of the species will become extinct.

\paragraph{Predator Death} 
Conversely, predators die based on a measure of hunger: the number of timesteps that have passed since the predator was born or since it last caught a prey. It is only possible for predators to die (death procedure) on certain timesteps, and the predator with the highest hunger is chosen to die. 

The timesteps in which the death procedure should occur are defined using an interval $\Delta$ (Eq.~\ref{eqd}): at timestep 0, a $\Delta$ is calculated based on the number of predators, and each next death procedure occurs after $\Delta$ timesteps. Before the death procedure starts, the measure of hunger is updated, and the $\Delta$ is updated after the death procedure ends. The $\Delta$ depends on the number of predators: the more predators, the lower the interval. This is meant to tackle overcrowding. Conversely, under-crowding is tackled because $\Delta$ also sets a limit for the death procedure to occur.

\begin{equation}
\Delta = 25 - p
 \label{eqd}
\end{equation}

where $p$ is the number of predators. This equation guarantees that the $\Delta$ is never below 2 (timesteps) because there must be a minimum of 7 prey in the population and therefore, a maximum of 23 predators.

%if a predator is at least within five units away

Additionally, there is an exception to dying from a high hunger measure: if a predator is at a minimum certain distance away from an observable prey, it will not die. In this case, the next oldest predator dies. This measure was implemented for situations when a predator might need just a little more time when it is very close to catching prey. Interestingly, if the prey being hunted suddenly switches tags, it could create a scenario where the predator instantly dies because it no longer falls within this exception.

\subsection{Metrics for system dynamics}

We utilize multiple metrics to analyze the system dynamics.

\paragraph{Attribution}
measures the performance of a species by calculating how much of the success or failure of the species can be attributed to selection pressure, as opposed to just an effect of randomness. The Attribution for the prey and predators is calculated differently.
 For prey, it means \textit{failure} in avoiding the predator; it is calculated through Eq.~\ref{sucprey}.

 \begin{equation}
a = \frac{p_i}{p_t}
 \label{sucprey}
\end{equation}
 
 where $p_i$ is the number of prey who were caught and had an inherited genotype, and $p_t$ is the total number of prey that were caught.
  
 For predators, it means \textit{success} in catching the prey; it is calculated through Eq.~\ref{sucpred}.

 \begin{equation}
a =  \frac{d_i}{d_t}
 \label{sucpred}
\end{equation}
 
 where $d_i$ is the number of predators who caught any prey and had an inherited genotype, and $d_t$ is the total number of predators that ever caught a prey.

\paragraph{Velocity}
This metric tracks whether a robot moves in a way to get closer (predator chases) or further away (prey evades) from its adversary. We measure the Velocity by first calculating the distance between the position of a robot  (P1) and the position of its adversary (A1). We verify if the robot moves closer or further away from the adversary position by checking the new position of the robot (P2) after 6 timesteps (12 seconds). If the distance between P2 and A1 is smaller than it was before, it means the robot moved in a way to get closer to its adversary and vice versa. The distance difference is then divided by the time delta, 12 seconds, to obtain the final value representing the Velocity (Eq. \ref{eqvelo}).

 \begin{equation}
 v=\frac{dist(P_1, A_1)-dist(P_2, A_1)}{12}
 \label{eqvelo}
 \end{equation}

\paragraph{Sticking to Walls Ratio} Due to the terrain being enclosed by four walls and the agents being unaware of their surroundings, robots may get stuck or move closely along the walls. We deem being stuck to a wall as being within one unit of distance to any of the four walls. For each timestep, we take the fraction of robots stuck to walls for each species. As an example, at time = 1250s, the prey had 10 out of 15 of their robots stuck to a wall. Therefore, their Stuck to Wall Ratio was $ 10/15$. 

\paragraph{Tag Symmetry}
The Tag Symmetry is of interest because it may support the existence of adaptive/reactive behavior. For instance, if predators have a consistently high tag average while the prey have a low tag average (or vice versa), this might mean that species are reacting to each other. For instance, the prey are trying to be invisible to the predator, so their tag is on average different from the average of the predators. It is important to highlight that this metric is unable to isolate active behavior from system dynamics. For example, we do not know if a certain value for this metric is due to the prey trying to be invisible or because all prey visible to the predator have been captured. 

The average tag is the value calculated from either prey or predators by considering the mean value of their tags. For example, if we had four predators with their tags (-1,1,1,1), the tag ratio would be $(-1+1+1+1)/4 = 0.75$. The more positive this value, the more individuals are tagged as +1.

The Tag Symmetry is calculated in every timestep by summing the average tags of the two species (including only inherited genotypes), e.g., for tag averages 0.75 and -0.71: $|0.75 + (-0.71)| = 0.04$. Values close to zero indicate higher symmetry. To establish a baseline, we contrast this symmetry score with a score obtained from 100,000 pairs of values uniformly distributed at random. This yielded a symmetry score of 0.66, equivalent to the average distance between two points on a line segment with a length of L=2 (from -1 to 1), which is L/3.

\section{Results}

\subsection{Attribution}
\begin{figure}[htbp]
\centerline{\includegraphics[scale=0.4]{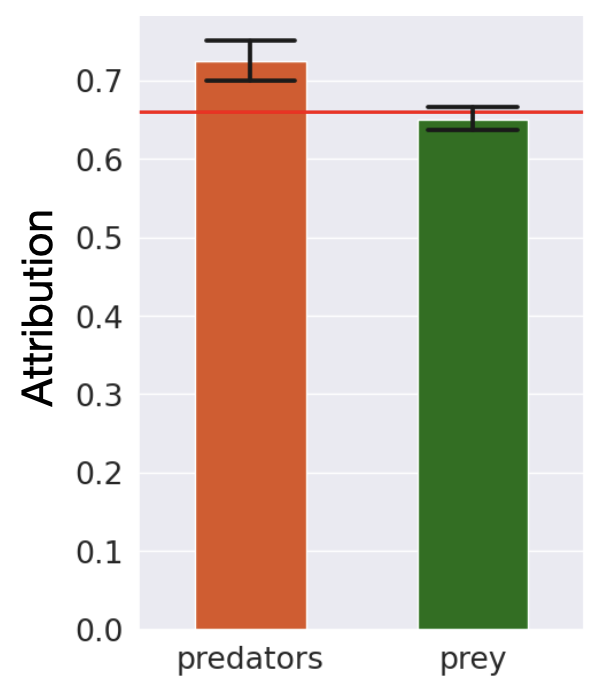}}
\caption{Differences in Attribution compared to a random baseline: the red line represents the random baseline. Higher values are better for predators, while lower values are better for prey.}
\label{figatr}
\end{figure}

This performance metric produces somewhat surprising results. Each robot, generated either through the random search or reproduction, could succeed (in catching or evading) by chance. When they succeed, they have a 66\% chance of reproducing, while there is no guarantee that their offspring will be as successful as them. Therefore, even if only 66\% of the individuals of a species succeeded, this could have been by chance - without any evidence that selection pressure took place towards a lineage of successful individuals. Therefore, we utilize 66\% as our baseline. This baseline is shown as the red line in Fig.~\ref{figatr}. To present increased performance, the predators would need to score higher than this baseline, while the prey would have to score lower. In the case of the predators, they have an average Attribution of 0.725. Therefore, the predators outperform the randomized solutions. On the contrary, prey performed indifferently to random solutions, with an Attribution of 0.65.  

\subsection{Velocity}
%There is a very divisive difference in the emergent average velocities between prey and predators. This is to be expected: 

\begin{figure*}
    \centering
    \includegraphics[scale=0.6]{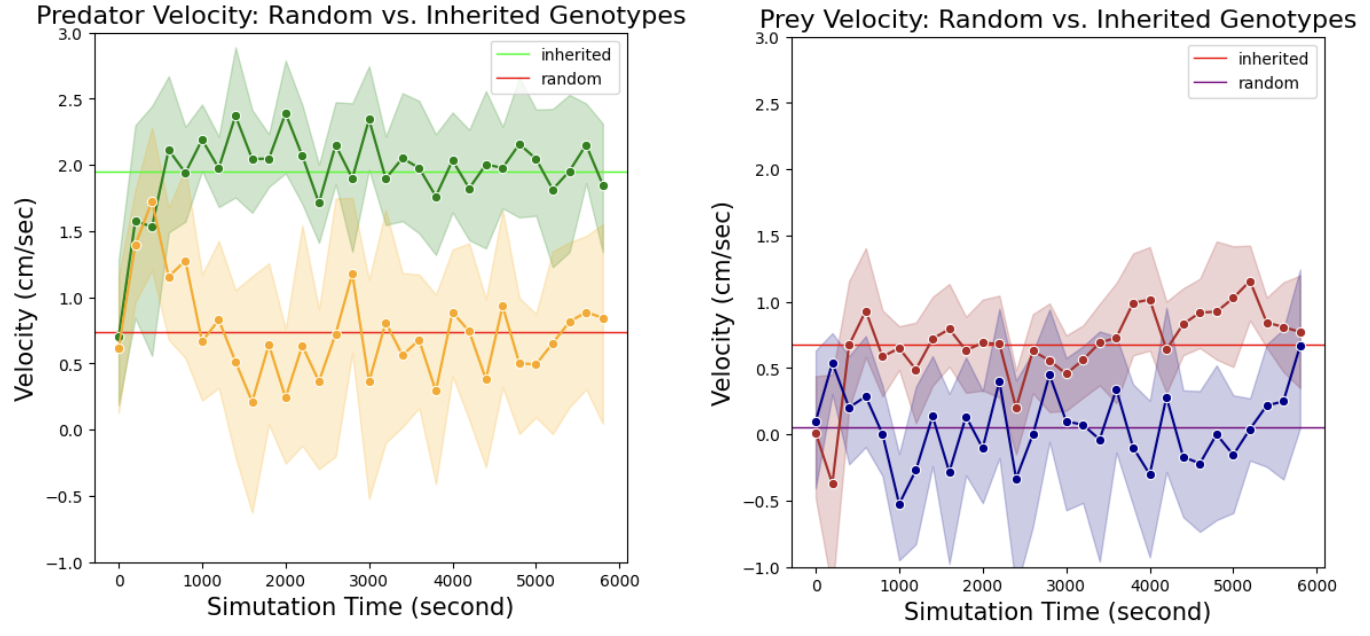}
    \caption{Progression of Velocity (toward or from adversary) across time: curves divide populations between randomly sampled robots and robots that reproduced (inherited).  The points in the curves are population averages measured every 200 seconds, resulting in 30 points. Straight lines represent averages of all points for each curve.}
    \label{velo}
\end{figure*}

A very effective strategy for predators is to move faster toward their prey and for prey to move faster away from predators. While it was expected that both species would have evolved to become better at chasing or avoiding each other, this happened much more successfully for the predators: predators move faster than prey towards the expected direction (Fig.~\ref{velo}). Prey move on average at 0.67 cm/s from their adversaries, while predators approach their adversaries on average at 1.95 cm/s - nearly three times. Furthermore, the evolving predators (generated through reproduction) become better than randomly generated predators after less than 1000 seconds and maintain this superiority until the end. On the other hand, the prey is not better than random at multiple eime

is a, no better than random. 

Note that although the targeted steering is pre-evolved, a better cognitive brain can, to some extent, increase velocity through a more assertive angle control towards the expected direction.

%We also observe that these behaviors are learned gradually. Initially, both populations have a small average velocity. This suggests that on average there is not a bias to either move away or towards their adversaries.  

\subsection{Sticking to Walls Ratio}

\begin{figure}[htbp]
    \centering
    \includegraphics[scale=0.37]{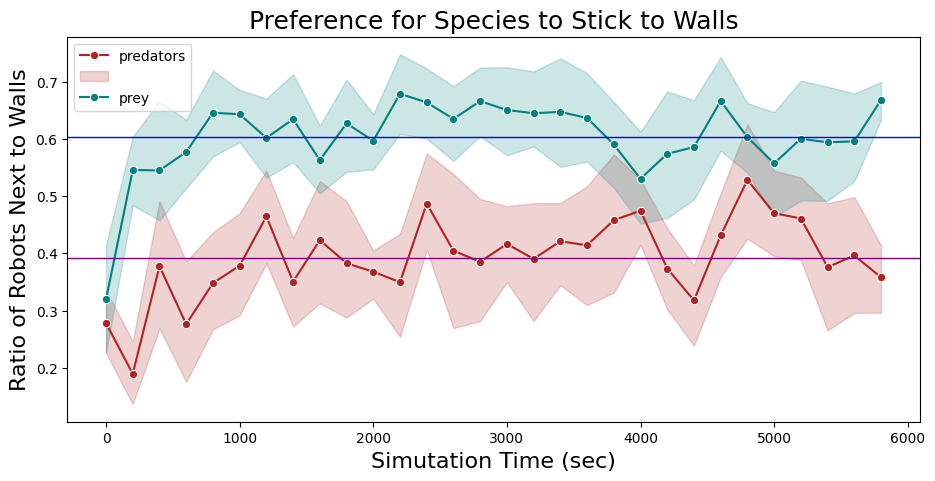}
    \caption{Progression Sticking to Walls Ratio across time: prey stick more to walls than predator. Straight lines represent averages of all points for each curve.}
    \label{sticking}
\end{figure}

Generally, prey favored sticking to nearby walls more than predators do (Fig. \ref{sticking}). The average ratio for predators, marked by the purple line, is 0.39. On the other hand, the average ratio for prey, marked by the blue line, is 0.60. This is about 50\% more.
 
While the discrepancy in their Sticking to Walls Ratio is a concrete phenomenon, it is unclear whether this behavior is beneficial. Perhaps the difference in behavior is merely a downstream consequence of: predators learn to track down prey; the prey end up at the wall because they are trying to evade; the prey remain unable to dodge the wall. 

\subsection{Tag Symmetry}
\begin{figure}[!h]
\centerline{\includegraphics[scale=0.30]{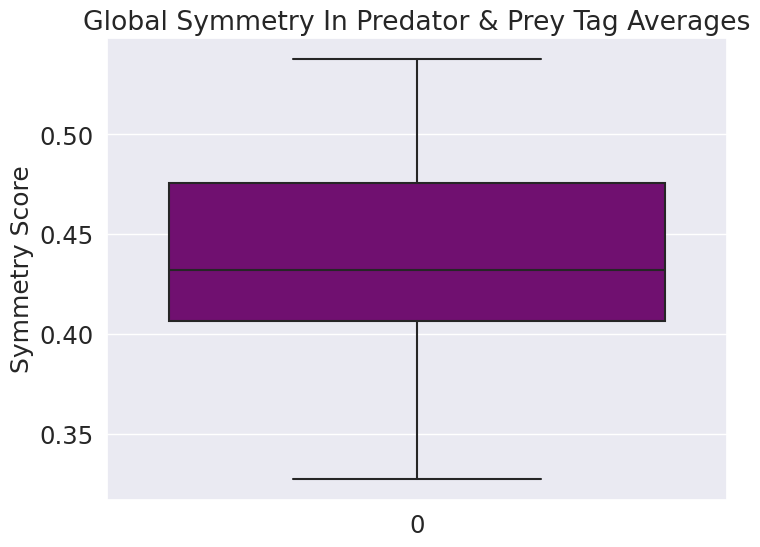}}
\caption{Distribution of Tag Symmetry averages among the independent experiments.}
\label{sym}
\end{figure}

The Tag Symmetry distribution across all ten runs is shown in Fig. \ref{sym}. The obtained average tag symmetry is 0.43, which is lower (65\%) than the random baseline of 0.66. Additionally, the minimum value is 0.33 and the maximum is 0.53, so that 0.66 falls completely outside the range: we can confidently conclude that the tag averages are more symmetric than random. However, an average of 0.43 is still not very high. Therefore, we do not use this as evidence to support the idea that species are reacting to each other.

\section{Conclusion}

This work has demonstrated how open-ended evolution can take place in a predator-prey scenario using modular robots. We presented evidence to support the emergence of evolved behavior beneficial to the survival of a species: the predators evolved towards higher effectiveness in capturing the prey. This was achieved without directly appealing to explicit selection mechanisms. On the other hand, despite the adaptive process of predators having been supported by clear evidence, the same did not occur to the prey. The ability of the prey to evade the predator was not significantly better than random. 

Notably, evading or chasing an adversary requires multiple partial behaviors, e.g., changing tags duly and moving away/toward the adversary (velocity). Therefore, it is possible that a species fails in accomplishing the behavior as a whole but succeeds in accomplishing sub-behaviors. Nevertheless, while the prey species presented some evidence of behavioral improvement regarding their velocity in evading the predator, this improvement was half of the time not significantly better than random.

One possible explanation for this shortcoming is the reproduction criterion utilized by the prey. To recapitulate, although there was no explicit goal applied through any selection mechanisms, the reproduction of the predators was conditioned to an explicit behavior: catching the prey. The reproduction of the prey, on the other hand, depended on an implicit behavior: being close to a predator when this one happens to die. These implicit versus explicit behaviors might have created different levels of selection pressure so that there was more pressure for the predators to improve than for the prey to improve: it is hard to determine if a prey reproduced because it had the ability to stay close enough to a predator without being caught, or if it was close enough to predator because it was unable to evade it. Additionally, the lack of an aging process for prey death might have influenced prey adaptation by creating less selection pressure for the prey.

%What seems to have happened instead is conventional evolution in a changing environment - in which the prey changes the environment of the predator through (mostly) random search. Thus, it can not be claimed that co-evolution occurred.
At this point, it is important to delineate two relevant concepts: reactive behaviors - the behavior of a species A changes in reaction to a change in the behavior of a species B, but without the behavior of A necessarily becoming superior/dominant to the behavior B; and co-evolution: there is an arms-race in which behaviors of species A and B become alternately superior to each other. Critically, while the predators did improve their success, the prey did not improve comparably, and therefore, we can not claim that co-evolution was achieved. As for reactive behavior, co-evolution is not necessarily required for it to occur. Nonetheless, the Tag Symmetry metric, whose purpose was to explore whether reactive behavior occurred due to symmetry, did also not result in convincing evidence.

Future work should explore more pressure-creating conditions for prey reproduction through a reproduction criterion defined by explicit behaviors. This is expected to promote prey success and foster an arms-race. Furthermore, the locomotion abilities in the current experiments were pre-evolved, and not a result of OEE: future work should also include targeted locomotion as a behavior subject to open-ended emergence. Finally, while the current experiments used a fixed modular morphology, future experiments should allow morphological evolution.

To conclude, we have presented evidence that the initial hypothesis regarding \textit{minimum criterion} is true in the current system: considering that there are no explicit selection mechanisms, the emergence of OEE depended on including an explicit behavioral criterion to allow reproduction. At the same time, the experimental setup does not allow isolating the effects of reproduction mechanisms from the lack of a prey aging process. 

% The relevance of the explicit criterion for reproduction was corroborated by the predators having adapted better than the prey despite the extra mechanism that reduced predator selection pressure: the rule that guaranteed the prevention of predator extinction at the cost of slaying prey.

%To conclude, the challenge to create conditions for open-ended evolution without explicit behavioral criterion for reproduction remains open, as it has been suggested before through the concept of \textit{minimum criterion}.

%While these observations would remain true in all other scenarios, this result is in line with the literature, lalalalal 

%Finally, it would be interesting designing more metrics to capture complex system dynamics that might not have been captured by our metrics. 

\bibliographystyle{IEEEtran}
\bibliography{oee}

\end{document}